\documentclass{article}

\usepackage[utf8]{inputenc}
\usepackage{lipsum}
\usepackage[ruled,vlined,linesnumbered]{algorithm2e}
\usepackage{graphicx}
\usepackage{float}
\usepackage{amsmath}
\usepackage{url}
\usepackage{microtype}
\usepackage{subfigure}
\usepackage{booktabs}

\usepackage{subfig}
\usepackage[accepted]{icml2020}

\newcommand\independent{\protect\mathpalette{\protect\independenT}{\perp}}
\def\independenT#1#2{\mathrel{\rlap{$#1#2$}\mkern2mu{#1#2}}}

\SetCommentSty{mycommfont}

\icmltitlerunning{Legally grounded fairness objectives}

\begin{document}

\twocolumn[
\icmltitle{Legally grounded fairness objectives}

\begin{icmlauthorlist}
    \icmlauthor{Dylan Holden-Sim\qquad}{}
    \icmlauthor{Gavin Leech\qquad}{}
    \icmlauthor{Laurence Aitchison}{}
\end{icmlauthorlist}
\begin{center}
    University of Bristol,\\
    \texttt{dh17830@bristol.ac.uk}
\end{center}


\icmlcorrespondingauthor{Dylan Holden-Sim}{dh17830@bristol.ac.uk}

\icmlkeywords{Machine Learning, algorithmic fairness}

\vskip 0.3in
]

\date{June 2020}

\begin{abstract}
    Recent work has identified a number of formally incompatible operational measures for the unfairness of a machine learning (ML) system.
    As these measures all capture intuitively desirable aspects of a fair system, choosing ``the one true'' measure is not possible, and instead a reasonable approach is to minimize a weighted combination of measures.
    However, this simply raises the question of how to choose the weights.
    Here, we formulate Legally Grounded Fairness Objectives (LGFO), which uses signals from the legal system to non-arbitrarily measure the social cost of a specific degree of unfairness.
    The LGFO is the expected damages under a putative lawsuit that might be awarded to those who were wrongly classified, in the sense that the ML system made a decision different to that which would have be made under the court's preferred measure.
    Notably, the two quantities necessary to compute the LGFO, the court's preferences about fairness measures, and the expected damages, are unknown but well-defined, and can be estimated by legal advice.
    Further, as the damages awarded by the legal system are designed to measure and compensate for the harm caused to an individual by an unfair classification, the LGFO aligns closely with society's estimate of the social cost.
\end{abstract}

\section{Introduction}

Automated decision making systems have not only become more prevalent, but are also being applied in increasingly sensitive contexts \cite{ClassifyingWithoutDiscriminating,KamishimaFairnessAwareRegularizer,cabitza2017unintended,military19}. This has led to demand for more transparent systems, as well as tools for assuring fairness. A key result here is that systems can discriminate on protected characteristics such as race, religion and gender, even when protected attributes are not an input to the system \cite{vzliobaite2016using,veale2017fairer}. 

The fundamental issue of algorithmic fairness is that no single definition of fairness captures the full phenomenon. Famously, the COMPAS recidivism prediction system was used in the criminal justice process in several US states. A ProPublica study argued that the COMPAS system was racially discriminatory, finding that African Americans labeled `High risk' were in fact 50\% less likely to reoffend than white defendants with the same label \cite{ProPublicaCompasAnalysis}. Flores et al responded by arguing that defendants with the same COMPAS recidivism score had approximately equal probability of recidivism \cite{NorthpointeDefence}. Here we see a direct clash between operationalisations of fairness.

Attempts to formalise fairness have yielded many such reasonable definitions \cite{EQOddsHardt, SolonFairMLBook, FairnessThroughAwareness}; ideally, we would fulfill these simultaneously. But we can prove that some sets of commonsensical definitions are incompatible, outside trivial cases \cite{InherentTradeoffs,BiasInRecidivismPrediction}.

Since perfect multi-measure fairness is almost always impossible, we instead aim at systems which minimise violations. 
But it is unclear which fairness definitions to relax and to what extent: subjective decisions about relative importance are required.
While it is clear that we cannot leave this task to the system implementors alone, the problem is actually far worse: as it is a question of values there may be irreconcilable differences between different individuals and there is no underlying well-defined but perhaps unknown ``correct answer''.
Without a well-defined correct answer even in principle, what weights should we pick?
We note that society can and must answer such question in other contexts using the legal system.
As such, we propose using the legal system to operationalise society's estimate of social costs.
The resulting Legally Grounded Fairness Objectives (LGFO) measure the damages awarded to those who were wrongly classified by the ML system under a putative lawsuit.

\subsection*{Minimising case cost as maximising social welfare}

Our solution is to find the classifier which minimises the damages to people classified differently under different definitions. That is, we minimise the \emph{legal} cost of choosing one measure over the other. This shifts the burden of selecting fairness measures onto the legal system and away from the programmer.

Our contention is that the social cost of an unfair classifier can be measured by the expected damages awarded to an individual given a false classification.

On first glance, setting the objective to minimised legal penalties looks inappropriate: as if privileging the interests of the system deployer. However, it is reasonable to view the size of legal damages is a proxy for social good, since: 
1) in principle, the law is designed to reflect the values of a society, including the broadest reading of fairness.
2) legal damages are intended to reimburse an individual for harm caused, as assessed by a judge. Thus, by minimising total damages (for instance by reducing unfairness and so the number of associated lawsuits) we simultaneously minimise harm.

A key advantage of this perspective is that we make use of the canonical process for balancing values and estimating social costs: the law, in this case civil law. While we should expect persistent disagreement about the nature of the social good, the legal system is the working mechanism society uses to approximate it, when informal means fail. 
In well-functioning jurisdictions, the legal process has a degree of public accountability, adaptiveness, and consensus - or anyway more than an average IT department \cite{burri,israni}. 

A second advantage is the relative availability of high-quality data. Our training signal is the monetary damages awarded to the plaintiff in algorithmic discrimination cases; in many jurisdictions, this data is openly available, e.g. \cite{LegalRulingDatabase}. We also need to elicit expert legal opinion on the type of fairness most applicable (or most often applied) in particular contexts, and on how much it would cost in a given case if a plaintiff's classification was changed. Given these, we can minimise a weighted combination of unfairness measures.




\subsection*{Related work}

$\mathrm{CFA}\theta$ \cite{CFATheta} is a fairness algorithm used to map distributions of raw scores towards the barycenter, a distribution occupying “middle ground” between the distributions of the different groups. The algorithm takes parameter $\theta$ which gives the degree of the mapping. $\mathrm{\theta} = 0$ leaves the raw scores unchanged, whereas $\mathrm{\theta} = 1$ sets all group distributions equal to the barycentre. It thus operationalises the tradeoff between individual fairness (low $\theta$) and group fairness (high $\theta$). A value $0 < \theta < 1$ corresponds to a partial mapping of group distributions towards this barycenter. $\theta$ is normative: its tuning would ideally be left to some democratic process. The issue is that selecting an appropriate $\theta$ requires a nuanced understanding of the algorithm by the decision maker, and the value is still a decision rather than calculated based on concrete values (i.e. legal costs).

The method proposed in \cite{FairnessThroughAwareness} encapsulates the idea that 'similar people should be treated similarly', a view known as Individual Fairness. This is achieved by enforcing a Lipschitz condition on the classifier:
For any two individuals $x, y$ at a distance $d(x, y) \in [0, 1]$ and map to distributions $M(x)$ and $M(y)$ respectively, the statistical distance between $M(x)$ and $M(y)$ is at most $d(x, y)$. Or
$D(M(x),M(y)) \leq d(x,y)$.

This is intuitive: the difference in group outcomes should be less than or equal to the difference in individuals.
This method is effective if domain knowledge can be used in constructing the distance function, i.e. if the normative work can be shared by other parties. But we have only shifted the subjectivity problem onto the distance function: whoever is given the task of defining $d$ still has to work in the absence of well-defined, unambiguous standards \cite{FairnessThroughComputationalBoundedAwareness}.

\subsection*{Our contributions}

We propose a new perspective in algorithmic fairness, using legal costs as a proxy for the social cost of a given fairness measure.

We define a method to account for multiple fairness measures and give an overall degree of unfairness, allowing for fairness maximisation.

We report experiments on a real-world dataset, showing that fairness measure combinations can correct naive correlations between the response variable and protected attributes. 

\section{Methods}
Our algorithm is a post-processing step for binary classifiers. We find the \textit{cost-minimal} decision boundary for each group: the pair $(t_0, t_1)$ where $t_i$ denotes the decision boundary (i.e. threshold value) for group $i$.

The fairness measures we use are initially binary properties: either satisfied perfectly or violated perfectly. To find decision boundaries which maximise a given fairness definition, we translate these notions into measures: functions of outcomes which are minimal at 0 (where the property is perfectly satisfied) and increase as we deviate further from the definition.

\subsection*{Unfairness as cost measure}
There are many proposed measures; here we focus on three, namely Sufficiency, Equalised Odds and Statistical Parity \cite{SolonFairMLBook, EQOddsHardt,FairnessThroughAwareness}. In \cite{InherentTradeoffs} it was proven that these measures are mutually incompatible outside of trivial cases - we cannot satisfy all three simultaneously (see Supplement, Proof 1). They include a positive and negative case, but we use only the positive case. Let $G$ denote the group of the defendants (here, a boolean for ethnicity), $Y$ be the ground truth label (here, actual recidivism risk category) and $\hat{Y}$ be the classifier's predicted label; let $1$ denote the high risk category and $0$ low risk.

The sufficiency of a classifier ($\mathrm{Suff}$) involves the difference in precision between groups (i.e. the probability of positive ground truth, given a positive prediction):\\
$\mathrm{Prec} = P(Y\!= 1 \ |\ \hat{Y}\!= 1)$ and

\vspace{-0.5cm}
\begin{align*}
    \mathrm{Suff} &= |\,\mathrm{Prec}_{\,G=0}
     - \mathrm{Prec}_{\,G=1} \,|
\end{align*}

Violation of $\mathrm{Suff}$ means that a positive prediction is more reliable for one group: and if positive classifications are less reliable for one group, then they cannot be used naively for decisions.

The Equalised Odds measure ($\Delta F$) involves the difference in false positive rate between groups: \\$\mathrm{FPR} = P(\hat{Y}\!= 1\ |\ Y\!= 0)$ and

\vspace{-0.5cm}
\begin{align*}
    \mathrm{\Delta F} &= |\, \mathrm{FPR}_{\,G=0}
         - \mathrm{FPR}_{\,G=1}  \,|
\end{align*}

Violating $\Delta F$ means we are more likely to wrongly predict that one group will reoffend than another group. This was the allegation in the ProPublica analysis: African Americans were more likely to be incorrectly labelled 'high risk' than white Americans \cite{ProPublicaCompasAnalysis}.

Finally, Statistical Parity ($\mathrm{SP}$) involves the difference between groups in the probability of predicting a positive label: 

\vspace{-0.5cm} 
\begin{align*}
    \mathrm{SP} &=  |\, P(\hat{Y}=1\ |\ G=0) - P(\hat{Y}=1\ |\ G=1) \,|
 \end{align*}

(Note that these are really \emph{unfairness} measures: that is, higher values indicate greater differences in handling different groups.)

\subsection*{Legally grounded fairness objectives}

Using any set of fairness measures $M$ and a set of example cases $X$, we can define the LGFO, the expected damages resulting from a hypothetical civil suit for wrongful classification:
\[
    \mathrm{LGFO} = \sum_{m \in M} P(m) 
                \sum_{x \in X} C(\hat{y}, y_m)
\]
where $\hat{y} = c(x)$ is the decision originally made by the ML system, $y_m$ is the decision that would have been made under fairness measure $m$, $P(m)$ is the probability that the court prefers that measure, and $C(y, y_m)$ is the misclassification cost of $y$ according to $m$.

\subsection*{The LGFO Algorithm}

Choose a particularly simple approach to minimizing the LGFO: we fix the classifer and modify group-dependent thresholds (Algorithm 1). 
This finds a separate score threshold for each group, such that the thresholds minimise overall multi-measure cost. 

Let the expected legal cost of changing an outcome from positive to negative be $\mathrm{P2N}$ and the cost of changing from negative to positive be $\mathrm{N2P}$. Let $\mathbf{X}$ be the set of all inputs to the classifier. Let $\hat{y}_i(x) \in \{0,1\}$ be the predicted label for $x$ after the raw score is thresholded by $s_i$, which is a tuple $(t_0,t_1)$ of per-group thresholds.

Let $P^*$ be a target number of positive classifications, which we set in order to avoid trivially fair cases (such as classifying all inputs as positive). Ideally we would consider cases where we exactly achieve $P^*$ positives; in practice this is not always possible.

Let $M$ be the set of fairness measures to balance, $\mathbf{C}_m$ be the set of costs incurred by applying measure $m$ alone over each threshold $s \in S$. The misclassification cost (of a threshold pair $s_i$ relative to the best threshold pair $s_j$) $O$ is, for example $x$:

\vspace{-0.5cm}
\begin{align*}
    O(x,s_i,s_j) = 
    \left\{
    \begin{array}{l}
      \mathrm{P2N},\quad \mathrm{if}\ \hat{y}_i(x) - \hat{y}_j(x) = 1\\
      \mathrm{N2P},\quad  \mathrm{if}\ \hat{y}_i(x) - \hat{y}_j(x) = -1\\
      0,\qquad  \mathrm{otherwise}
    \end{array}
    \right.
\end{align*}
\vspace{-0.2cm}

The output is the threshold pair which gives the minimum summed cost $C_{sum}$; that is, the lowest cost we can obtain under all measures.

\begin{algorithm}[!ht]
    \SetKwFunction{algo}{algo}\SetKwFunction{proc}{get\_thresholds}
   \caption{\, Minimizing LGFO}
   \KwIn{
   $M$ : set of fairness measures to balance, \\
   \qquad\quad $\mathbf{X}$: examples, \\
   \qquad\quad $\hat{\mathbf{Y}}$: classifier scores, $c(x) \,\, \forall x \in X$ \\
   \qquad\quad $P^*$: target number of positives
   }
    \vspace{0.3cm}
    
    $S$ = \proc{$\mathbf{X}, \hat{\mathbf{Y}}, P^*$}\;
    
    $\mathbf{C}_m = [\,]$ \;
    
    \vspace{5pt}
    
   \ForAll{$m \in M$} {
   
        $s_{\mathrm{min}} = \underset{s \in S}{\mathrm{argmin}}(m(s))$ \;
   
        \ForAll{$s_i \in S,\, s_i \neq s_\mathrm{min}$} {
   
        $C = \sum_{x \in X}{O(x,s_i,s_{\mathrm{min}})}\  $\; 
        
        $\mathbf{C}_m[s_i] = C$\;
    }
   }
   \vspace{2mm}
   \ForAll{$s_i \in S$} {
   $\mathbf{C}_\mathrm{sum}[s_i] = \sum_{m \in M}
   \mathbf{C}_m[s_i]$\;
   }
   \vspace{2pt}
   \Return{ $\underset{s \in S}{\mathrm{argmin}}(\mathbf{C}_\mathrm{sum} )$ } \;
   
   \vspace{5mm}
   \SetKwProg{myproc}{Procedure \,}{}{}
   \myproc{\proc{$\mathbf{X}, \hat{\mathbf{Y}}, P^*$}} {
    $S = [\,]$\;
    \vspace{1mm}
    
    \ForAll{$t \in [0, 0.02, ..., 1]$}{
        $n_{p'},\ n_{q'} = \infty$\;
        
        \ForAll{$t_0,\ t_1 \in [0, 0.02, ..., 1]$}{
        $s_p = (t,t_1)$\;
        
        $s_q = (t_0,t)$\; 
        
        $n_p = \sum_{x \in \mathbf{X}} \ \hat{y}_p(x) $ \;
            
        $n_q = \sum_{x \in \mathbf{X}} \ \hat{y}_q(x) $ \;
        
        \uIf{ $|n_p - P^*| < n_{p'}$}{
        
            $n_{p'} = n_p$\;
            
            $s_{p'} = s_p$\;
            
        }
        \uIf{ $|n_q - P^*| < n_{q'}$}{
        
            $n_{q'} = n_q$\;
            
            $s_{p'} = s_q$\;

        } 
        }
        append $s_{p'}$ to $S$\;
        
        append $s_{q'}$ to $S$\;
    }
        
        \KwRet $S$ \; 
    }
\label{lgfo}
\end{algorithm}
We validated LGFO using the COMPAS dataset \cite{AIF360,ProPublicaCompasAnalysis}: we implemented a PyTorch binary classifier predicting the probability of belonging to the 'High chance of violent recidivism' class. The original COMPAS system used a scoring system; our approach mirrors ProPublica in merging Medium and High risk categories and constructing a binary classifier for this new group.

\subsection{Cost sensitivity}

A property that then naturally arises is \textit{cost-sensitivity}. Consider a measure cost-sensitive if it leads to large changes in cost for small changes in absolute measure value.

Cost-insensitive measures can be relaxed to a much greater degree without incurring large social cost. This provides an opportunity to improve on other measures which are more sensitive to cost, leading to an output that is more fair under more definitions.

\section{Results}

\begin{figure}[ht]
\begin{center}
\centerline{\includegraphics[width=\columnwidth]{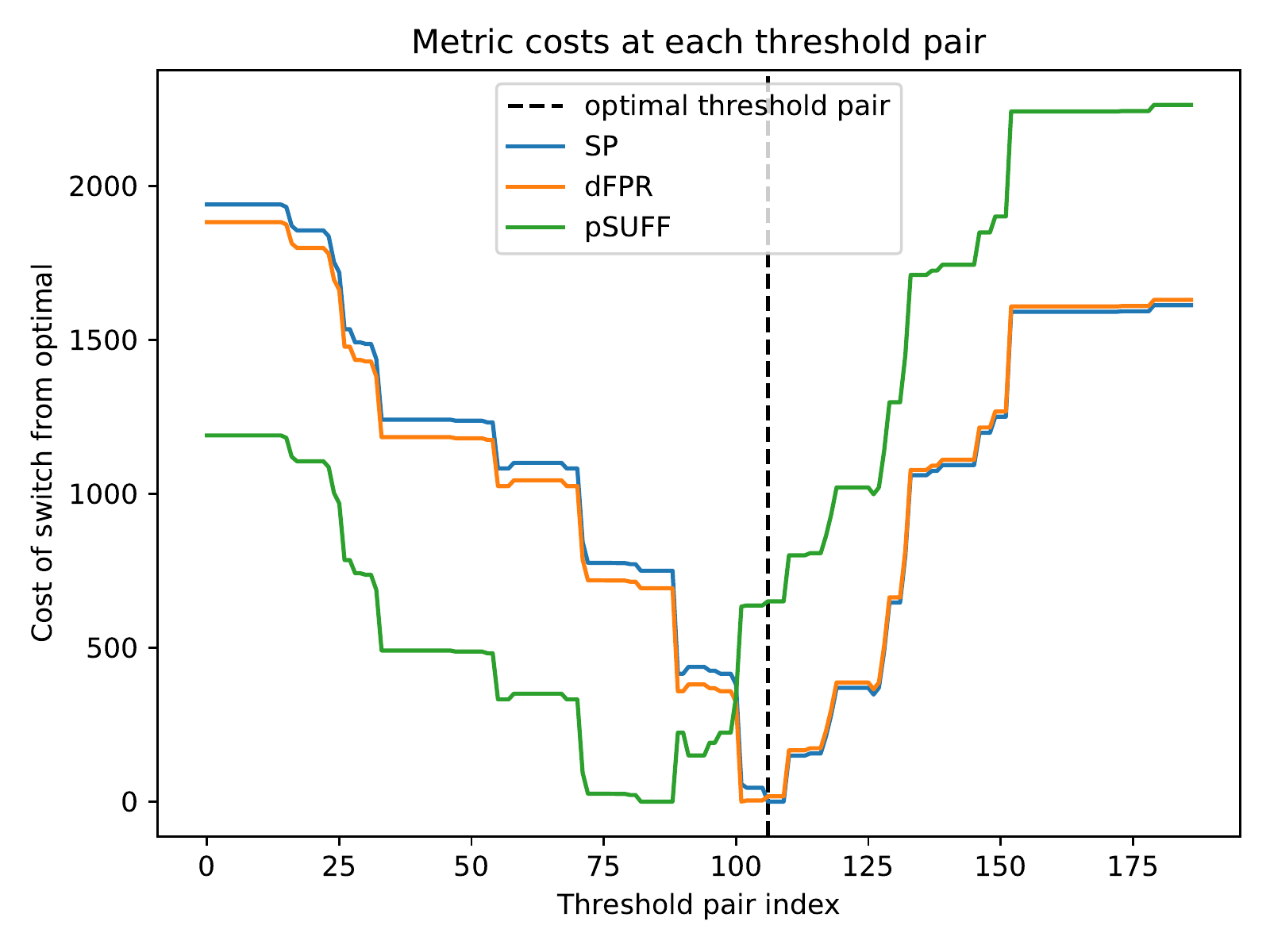}}
\caption{Costs of deviating from the optimal configuration of each measure in LGFO.}
\label{three measure costs}
\end{center}
\vskip -0.2in
\end{figure}

\begin{figure}[ht]
\begin{center}
\centerline{\includegraphics[width=\columnwidth]{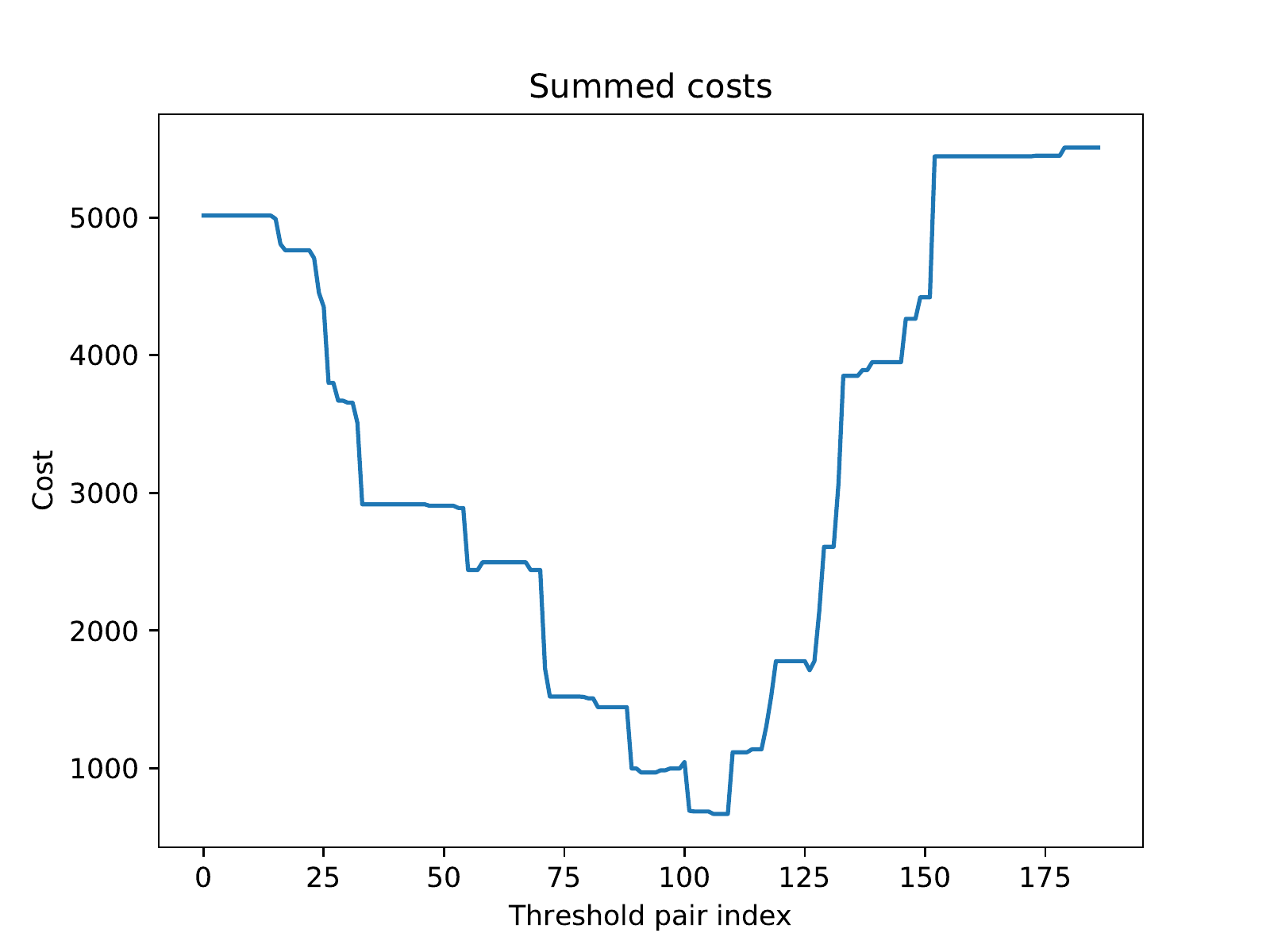}}
\caption{Summed individual measure costs. The minima is our cost-optimality configuration. This aligns with the optimal for $\Delta F$.}
\label{three measure cost sum}
\end{center}
\end{figure}

Figure \ref{three measure vals} compares the raw values of our chosen fairness measures at different thresholds; Figure \ref{three measure costs} shows the cost of violating the fairness measure. Figure \ref{three measure cost sum} then shows the summed cost which yields the minimal-cost fair configuration. 

It is helpful to visualise costs as a curve by ordering threshold pairs from highly preferential treatment for one group to highly preferential treatment for the other, with the midpoint being equal treatment for both groups. The 'Threshold pair index' then represents the index of this ordered collection.

\subsection*{Fairer COMPAS predictions}

\begin{figure}[ht]
\begin{center}
\centerline{\includegraphics[width=\columnwidth]{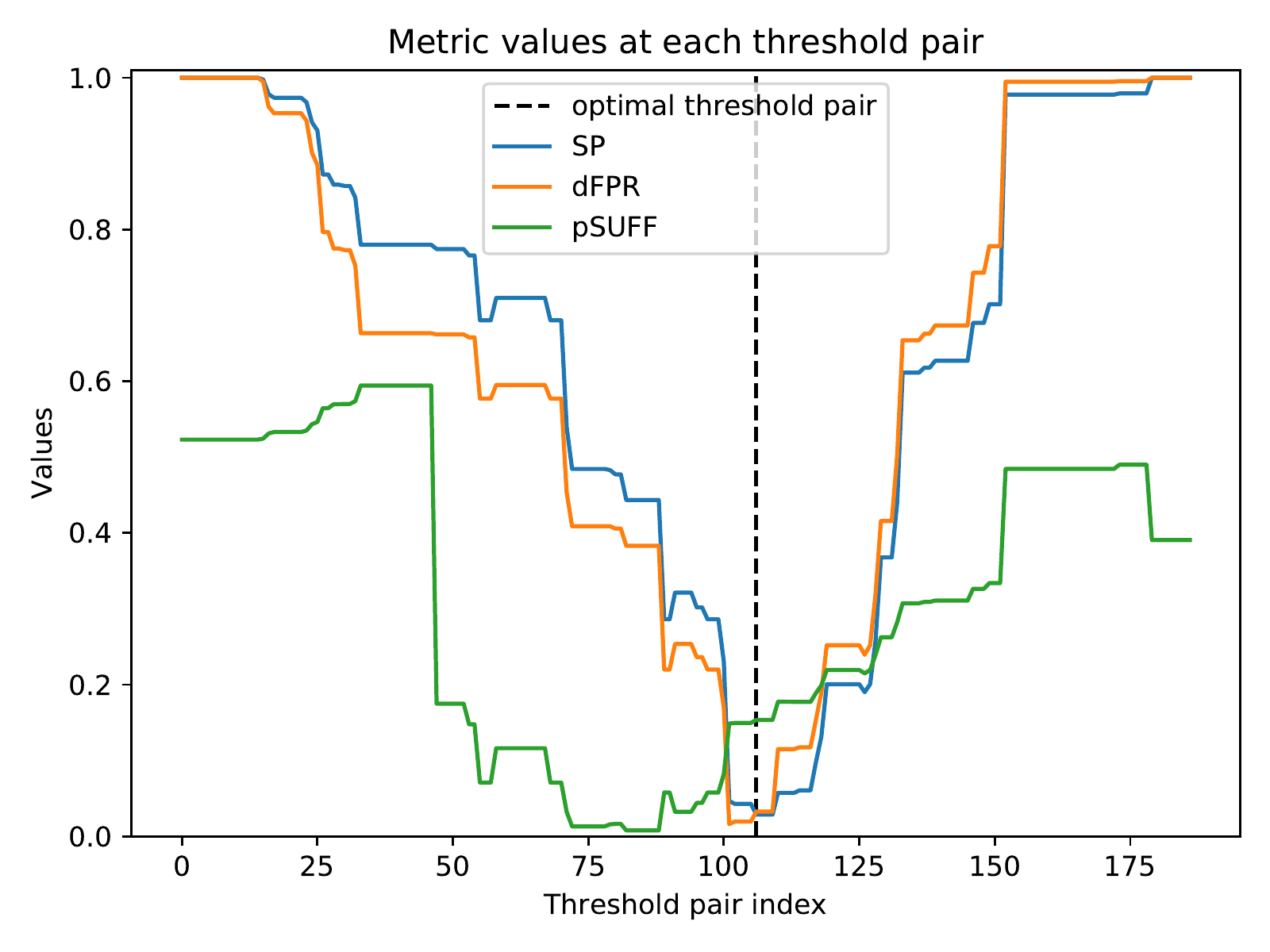}}
\caption{Unfairness values over threshold choices.}
\label{three measure vals}
\end{center}
\vskip -0.2in
\end{figure}

In Figure \ref{three measure vals} we see that $\mathrm{SP}$ and $\Delta F$ roughly agree on the fairest region. This is because both encapsulate a similar notion of fairness, penalising discrepancies between group outcomes. The cost-optimal thresholds found are [0.54,0.41], corresponding to slightly favourable treatment for African Americans.

\subsection*{LFGO vs uncorrected classification}
To evaluate our algorithm, we compare the LGFO classification to the uncorrected classification (equivalent to the threshold pair $[0.5, 0.5]$).

We see our corrected model makes the trade off of Sufficiency for Statistical Parity and Equalised Odds. There is also a small accuracy decrease of 2\%. Accuracy is maintained, since LGFO mostly changes classifications only for defendants receiving uncertain predictions. Inputs with predictive values close to 0 or 1 will only see changes to their outcomes in extreme decision boundaries, i.e. those that are unfair by our measures. Our algorithm only changes outcomes for a fraction of individuals, those the raw classifier is more likely to mislabel. 

\begin{figure}[t]
\begin{center}
\centerline{\includegraphics[width=\columnwidth]{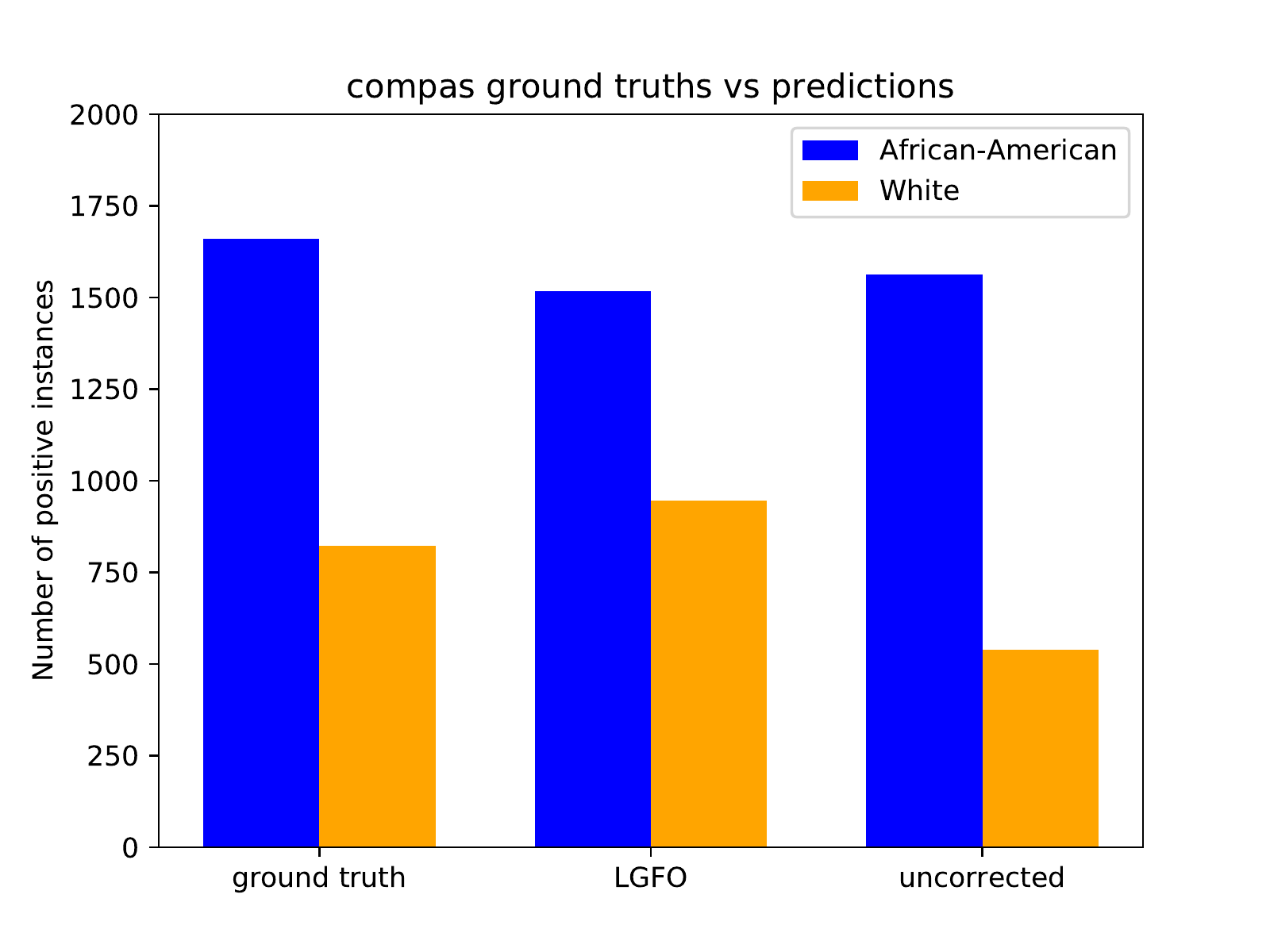}}
\caption{Comparing corrected and uncorrected per-group positive predictions to ground truth.}
\label{outcome comparison hist}
\end{center}
\vskip -0.2in
\end{figure}

Looking at Figure \ref{outcome comparison hist}, we see that the Uncorrected predictor significantly under-represents white plaintiffs in positive predictions versus the ground truth data. LGFO corrects for this, bringing the number of predictions closer to the ground truth.

\begin{table}[]
\caption{measure results for the uncorrected and LGFO classifiers}
\label{sample-table}
\begin{center}
\begin{small}
\begin{sc}
\begin{tabular}{lccc}
\toprule
Measure & Uncorrected & LGFO  \\
\midrule
Statistical Parity & $0.236$ & $\mathbf{0.029}$ \\
Suff & $\mathbf{0.062}$ & $0.154$ \\
$\Delta F$ & $0.162$ & $\mathbf{0.033}$  \\
Accuracy & $\mathbf{67.0\%}$ & $65.7\%$  \\
\bottomrule
\end{tabular}
\end{sc}
\end{small}
\end{center}
\vskip -0.1in
\end{table}

\subsection*{Cost-sensitivity}

We see that $\mathrm{Suff}$ is the most cost-sensitive measure: we see the same cost incurred when $\mathrm{Suff} = 0.2$ as we do when $\Delta F = 0.5$. This explains the result in Table \ref{sample-table}, in which Sufficiency actually decreases after applying LGFO: this is a principled trade for greatly reduced unfairness on the other measures. 
The key result is that inspecting the measure values themselves is insufficient to judge the actual relative fairness of two classifiers: taking the damages into account shows that lower measure fairness can occur when damages are reduced.

\subsection*{Illustrative scenarios}

We now investigate LFGO with counterfactual scenarios. 

\subsubsection*{Intermediate Fairness}

\begin{figure}[t]
\begin{center}
\centerline{\includegraphics[width=\columnwidth]{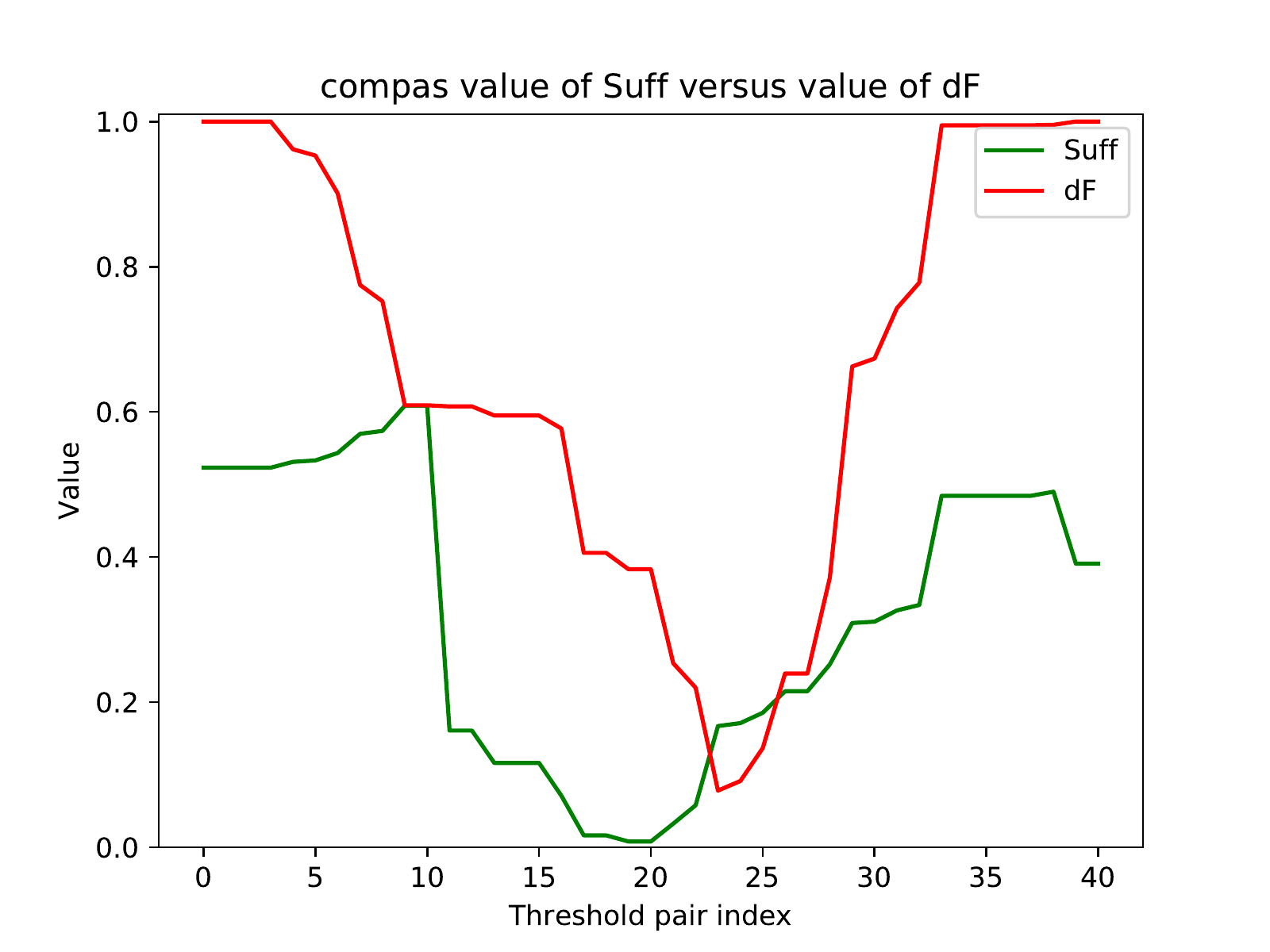}}
\caption{Value of the measures at each threshold pair.}
\label{measure vals}
\end{center}
\vskip -0.2in
\end{figure}

\begin{figure}[t]
\begin{center}
\centerline{\includegraphics[width=\columnwidth]{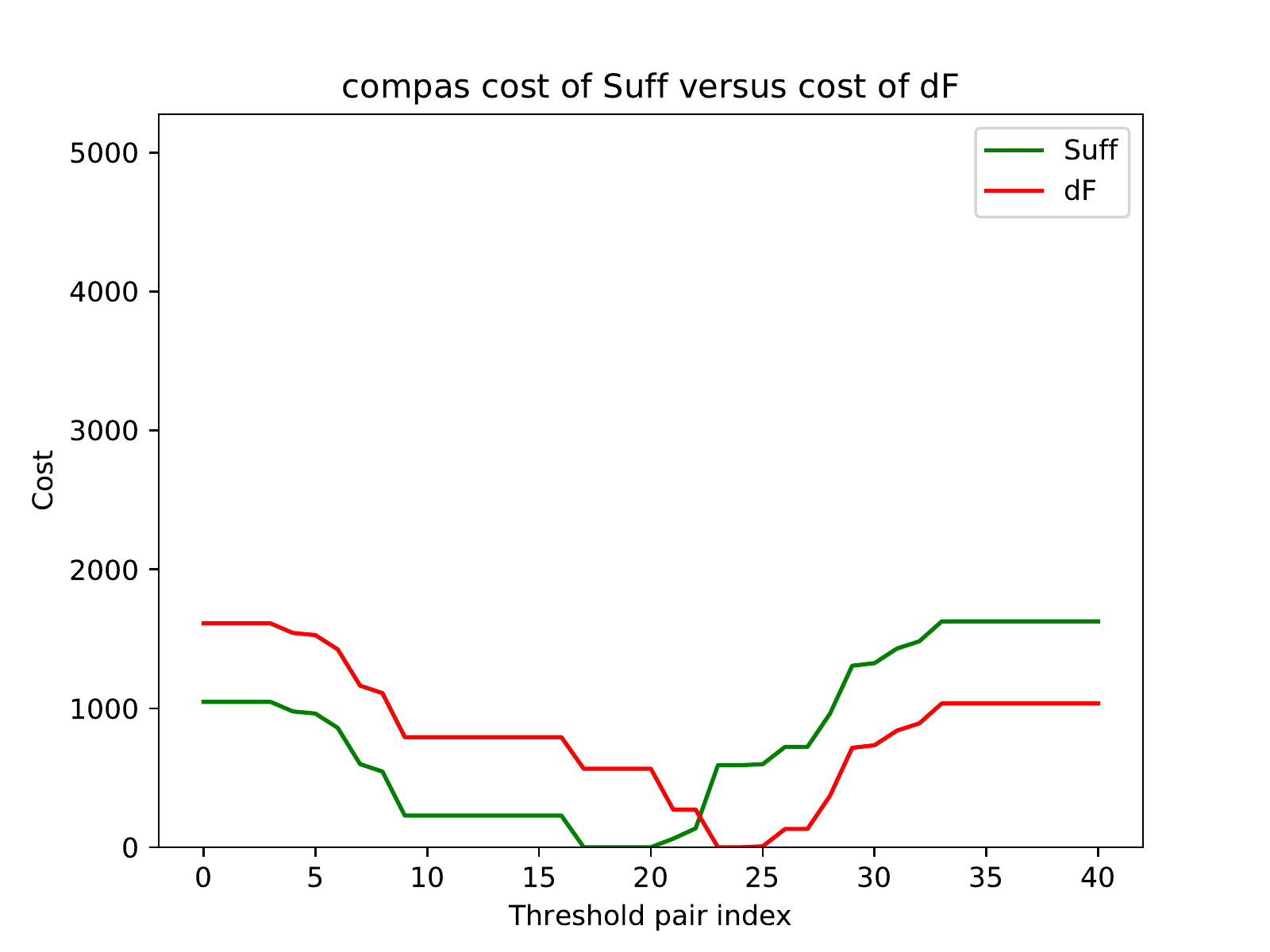}}
\caption{Costs of deviating from the optimal configuration for each measure.}
\label{measure costs}
\end{center}
\vskip -0.2in
\end{figure}

\begin{figure}[t]
\begin{center}
\centerline{\includegraphics[width=\columnwidth]{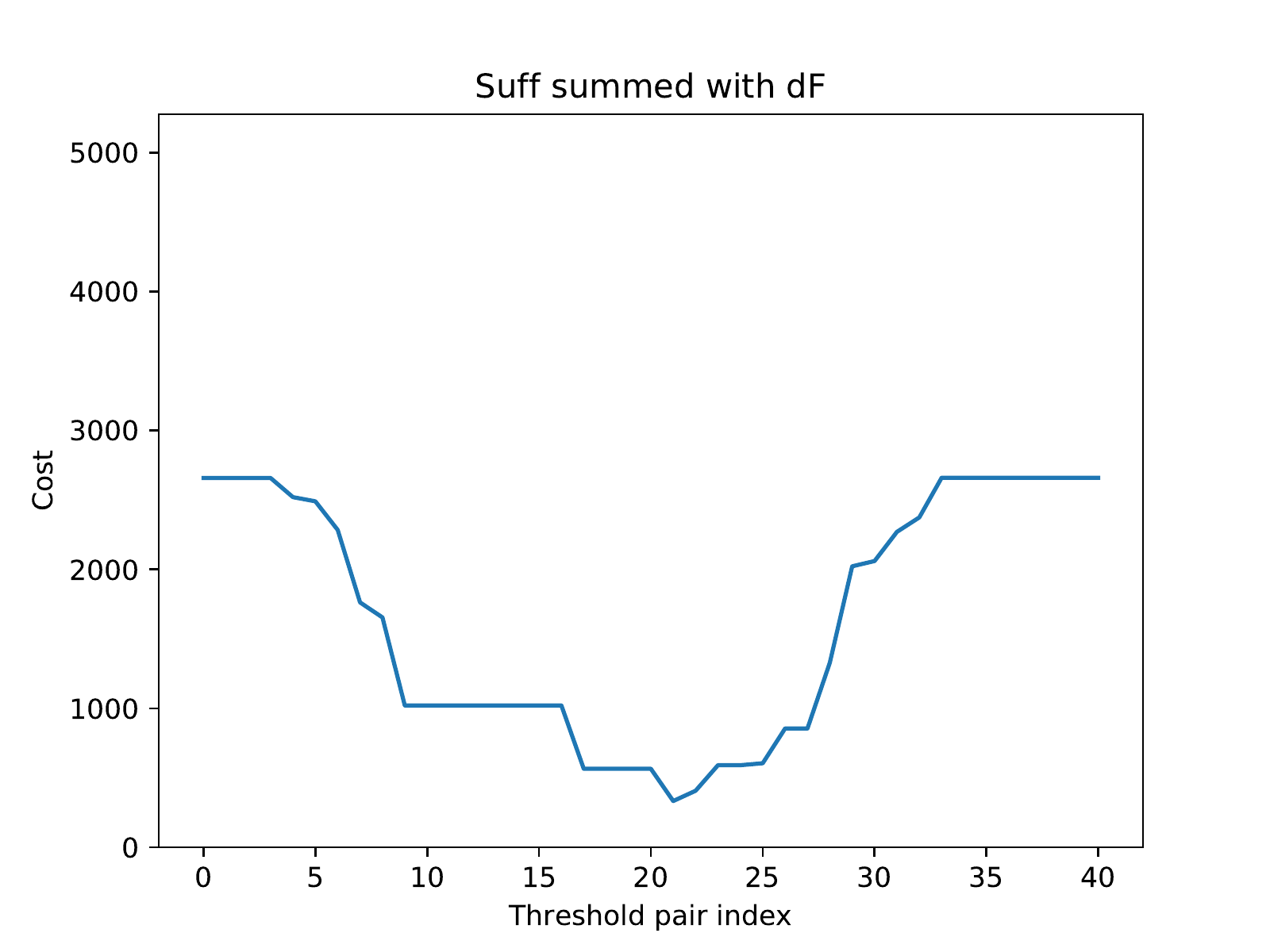}}
\caption{Summed individual measure costs. The minimum corresponds to our cost-optimal fair configuration - a trade off between both measures.}
\label{measure cost sum}
\end{center}
\vskip -0.5in
\end{figure}

In place of true legal costs, this scenario sets $\mathrm{P2N} = 0$ and $\mathrm{N2P} = 1$. This corresponds to the cost of changing a prediction to 'highly likely to reoffend' (a false positive) being higher than the converse (false negative).

Next compare the cost of violating a measure to the degree of violation ($C_m$ vs $m(X)$). In Figure \ref{measure vals} we see that to minimise $\mathrm{Suff}$ (around $x=20$), we incur high $\Delta F$; but conversely a very low $\Delta F$ (around $x=23$) results in moderate $\mathrm{Suff}$. When comparing the costs, however (Figure \ref{measure costs}), we see that costs are actually equal for both thresholds.

Figure \ref{measure cost sum} shows the minimum summed-cost point. This cost-minimal configuration occurs in-between the optimal $\mathrm{Suff}$ or $\Delta F$, corresponding to partial unfairness on both accounts; but this simultaneous relaxation yields a better social outcome than optimising for either alone.

\subsubsection*{Single-type Fairness}

\begin{figure}[!ht]
\vskip 0.2in
\begin{center}
\centerline{\includegraphics[width=0.99\columnwidth]{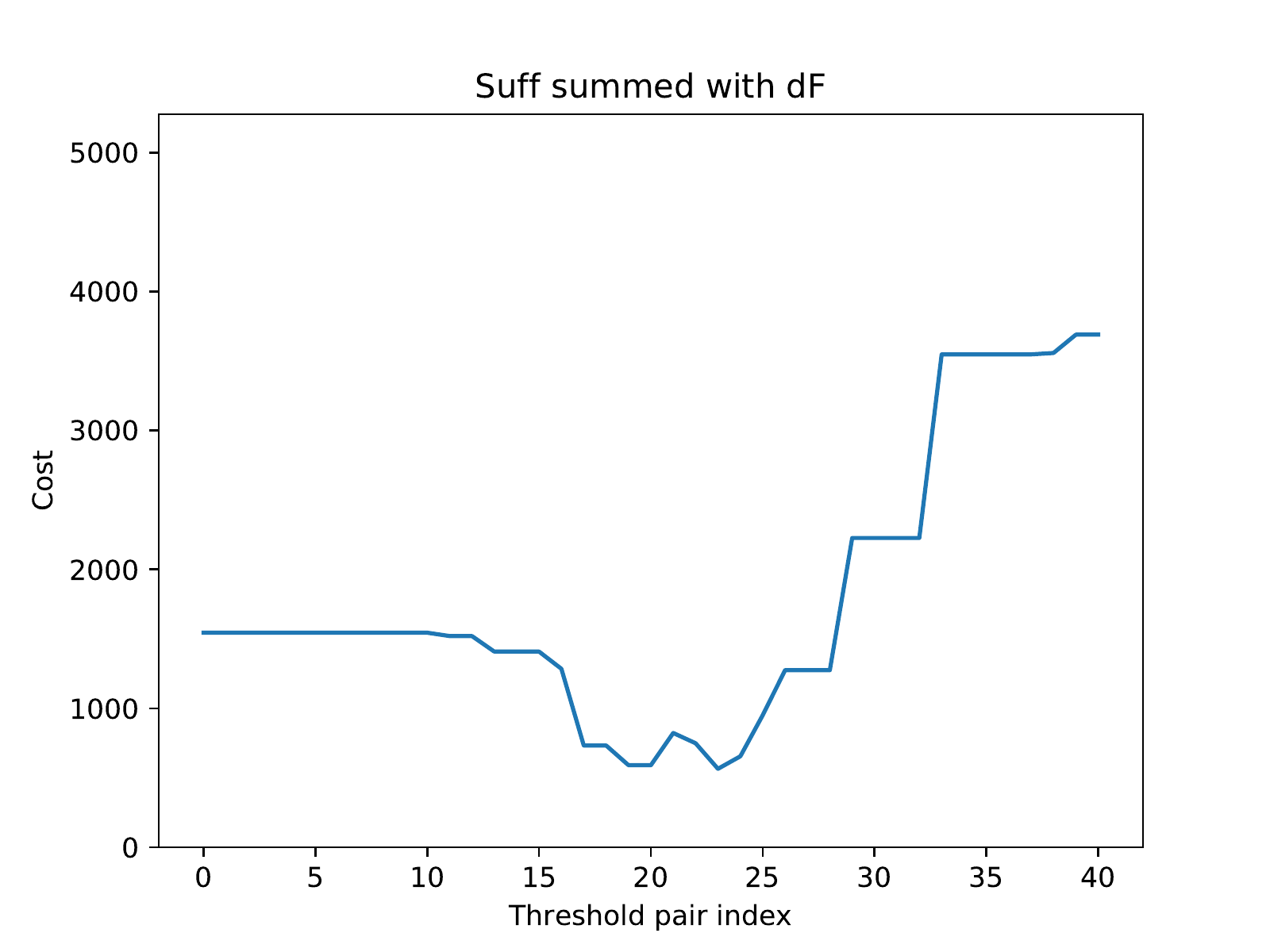}}
\caption{Summed individual measure costs. Here cost-optimality occurs under optimisation of a single measure. In this case partial satisfaction of both measures yields a higher cost.}
\label{no optimal measure cost sum}
\end{center}
\end{figure}

Figure \ref{no optimal measure cost sum} is from a scenario with $\mathrm{P2N} = 1$ and $\mathrm{N2P} = 0$. Intermediate thresholds yield higher cost than optimising for either measure individually. The local maximum occurs at the same location as the minimum in Figure \ref{measure cost sum}, which implies that while the measure values are reasonable, we actually cause more harm in trying to balance both than when a single measure is used.


\section{Discussion}

LGFO has several virtues: it brings fairness into business operations, by setting an unambiguous, hard-to-game monetary incentive towards fair systems. It also allows for stakeholders other than the technical team to contribute to the system design, and makes use of long-standing legal expertise on  decision-making in complex social situations.

We found that our algorithm was able to correct for erroneous bias in a neural classifier using a real-world dataset, while conserving performance. Fairness algorithms should remain performance-competitive, to make it more likely that they are actually implemented. On our dataset and classifier we noted a small (2\% relative) loss of overall accuracy from applying LGFO; however, as shown in Figure \ref{outcome comparison hist}, this minor performance cost impacts each group differently.

\subsection*{Limitations}
Our approach cannot be applied immediately to arbitrary data, but requires a careful elicitation step. This is due to the lack of explicit use of formal fairness measures in most legal systems \cite{XiangLegalCompatibility}.

In the present experiments, we use ProPublica's binarised risk groups, which does not reflect the actual use of deployed systems, nor the finer-grained information in the original scores.


We only handle binary classification; however, extensions of the method to other learning settings is a possible area for future experimentation. We could also extend the method to other stages of the ML pipeline. The LGFO value could, for instance, be used in the training stage of the classifier.

Clearly, the legal system is also an imperfect estimator of social cost, and has its own biases \cite{burden,berrey2012situated}. But it seems unlikely to be more biased (or unaccountable) than a lone technical team with no clear incentives towards fairness. LGFO ties ML into an existing democratic process with greater domain knowledge of the tradeoffs involved.


The LGFO estimation process is unavoidably local: the distribution over fairness definitions, and the damages involved, will vary greatly between jurisdictions. But this is just the converse of the method's strength: that it makes use of actual domain knowledge.

\newpage
\bibliography{references}
\bibliographystyle{icml2020}

\newpage
\onecolumn

\appendix

\section{Proofs of metric incompatibility}
 
The proofs rely on the following three short theorems from \cite{AllOfStatistics} page 264 and follow the methods of \cite{SolonFairMLBook}. Also see \cite{FriedlerImpossibleFairness} for proof of the incompatibility of the fairness metrics.

\begin{equation}
X \independent{} Y | Z \implies Y \independent{} X | Z\\
\end{equation}
\begin{equation}
X \independent{} Y | Z\ \mathrm{and}\ X \independent{} Z | Y \implies X \independent{} (Y,Z)
\end{equation}
\begin{center}
Finally, assuming all events have positive probabilities:
\end{center}
\begin{equation}
 X \independent{} Y | Z\ \mathrm{and}\ X \independent{} Z \implies X \independent{} (Y,Z)\ 
\end{equation}

We give the probabilistic definitions for each of these three metrics here.
\begin{itemize}
    \item Equalised Odds: $\hat{Y}\independent{}A|Y$\\
    \item Statistical Parity: $\hat{Y}\independent{}A$\\
    \item Sufficiency: $Y \independent{} A | \hat{Y}$\\
\end{itemize}

\subsection{Statistical Parity vs Sufficiency}

Assuming Statistical Parity ($\hat{Y} \independent{A}$) and Sufficiency ($Y \independent{A|\hat{Y}} \equiv A \independent{Y|\hat{Y}}$) (equivalence from theorem (1)). Then by theorem (3), we have $A\independent{(Y,\hat{Y})}$. This means that enforcing statistical parity and sufficiency simultaneously only holds when $A \independent{Y}$, meaning when base rates across groups are equal. This, of course, is unrealistic with a real world dataset.

\subsection{Equalised Odds vs Sufficiency}

Given Equalised Odds ($\hat{Y}\independent{}A|Y$) and Sufficiency ($Y \independent{A|\hat{Y}}$). Rearranging both using theorem (1) gives $A\independent{}\hat{Y}|Y$ and $A \independent{Y|\hat{Y}}$. Then from theorem (2) we have $A \independent{(Y,\hat{Y})}$ which is identical to the problems in Statistical Parity vs Sufficiency.


\section{Proof of incompatibility of Statistical Parity and Equalised Odds in the binary case.}

Follows \cite{SolonFairMLBook} page 55 prop. 3.

Assume $Y$ is binary, $A$ is not independent of $Y$, and $\hat{Y}$ is not independent of $Y$. Then, independence and separation cannot both hold.
Assume $Y \in \{0, 1\}$. In its contra-positive form, the statement we need to show is
$A \independent \hat{Y}\ and\ A \independent \hat{Y}|Y \implies A \independent Y$ or $\hat{Y} \independent Y$.

By the law of total probability\footnote{$$ P(A\mid C)=\sum _{n}P(A\mid C\cap B_{n})P(B_{n}\mid C)$$}, $$P(\hat{Y} = \hat{y} | A = a) = \sum{P(\hat{Y} = \hat{y} | A = a,Y = y)P(Y = y | A = a)}$$

Applying the assumption $A  \independent  \hat{Y}$ and $A  \independent  \hat{Y} | Y$, 
this equation simplifies to $$P(\hat{Y} = \hat{y}) = \sum{P(\hat{Y} = \hat{y} | Y = y)P(Y = y | A = a)}$$

Applied differently, the law of total probability\footnote{$$ P(A)=\sum _{n}P(A\mid B_{n})P(B_{n})$$} also gives
$$P(\hat{Y} = \hat{y}) = \sum{P(\hat{Y} = \hat{y} | Y = y)P(Y = y)}$$

Combining this with the previous equation, we have
$$\sum{ P(\hat{Y} = \hat{y} | Y = y)P(Y = y) } = \sum{P(\hat{Y} = \hat{y} | Y = y)P(Y = y | A = a) }$$

Inspection of this equation reveals that when $y$ is binary, this equation can only be satisfied if $A  \independent  Y$ or $\hat{Y}  \independent  Y$.

We can rewrite the equation more compactly using $$p=P(Y=0),\ p_a = P(Y=0|A=a),\ \hat{y}_y =P(\hat{Y}=\hat{y}|Y=y) $$ as:
$$p\hat{y}_0 +(1-p)\hat{y}_1 = p_a \hat{y}_0 + (1 - p_a) \hat{y}_1$$

Subtracting $\hat{y}_1$ from both sides gives: $$p*(\hat{y}_0 - \hat{y}_1) = p_a * (\hat{y}_0 - \hat{y}_1)$$
This equation can only be satisfied if $\hat{y}_0 = \hat{y}_1$, in which case $\hat{Y}  \independent  Y$, or if $\forall a, p=p_a$, in which case $Y \independent A$

The problem of $A \independent{Y}$ is discussed in the previous section. $\hat{Y} \independent{Y}$ is problematic as it means our classifier does not have utility - predictions are independent of the true values. 

\end{document}